%% file: main.tex
\documentclass[letterpaper, 10 pt, conference]{ieeeconf}  

\IEEEoverridecommandlockouts                              

\usepackage{float}
\usepackage{graphicx}
\usepackage{caption} 
\usepackage{overpic}
\usepackage{amsmath}
\usepackage{booktabs}
\usepackage{tabularx}

\title{\LARGE \bf
LLGS: Unsupervised Gaussian Splatting for Image Enhancement and Reconstruction in Pure Dark Environment
}

\author{Haoran Wang$^{1*}$, Jingwei Huang$^{2*}$, Lu Yang$^{2}$, Tianchen Deng$^{3}$, Gaojing Zhang$^{1}$, and Mingrui Li$^{4\dagger}$
\thanks{*Euqal Contribution, $^{\dagger}$Corresponding author.}%
\thanks{$^{1}$Haoran Wang and Gaojing Zhang are with the School of Engineering and Informatics, University of Sussex}%
\thanks{$^{2}$Jingwei Huang and Lu Yang are with the Department of Automation Engineering, University of Electronic Science and Technology of China}%
\thanks{$^{3}$Tianchen Deng is with the Department of Automation, Shanghai Jiao Tong University}%
\thanks{$^{4\dagger}$Mingrui Li is with the Department of Computer Science, Dalian University of Technology
        {\tt\small mingruili@mail.dlut.edu.c}}%
}

\begin{document}

\maketitle
\thispagestyle{empty}
\pagestyle{empty}


\input{secs/0_abstract}
\input{secs/1_introduction}
\input{secs/2_relatedwork}
\begin{figure*}[htbp!]
    \centering
    \includegraphics[width=\linewidth]{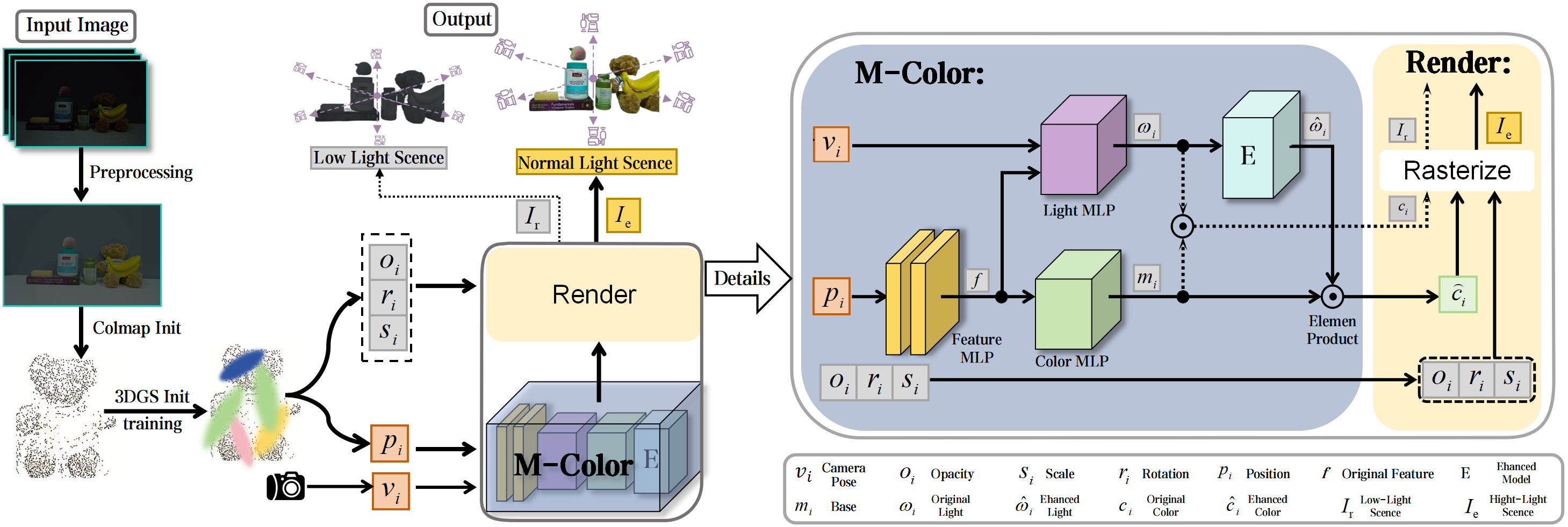}
    \caption{\label{fig:overview}Pipeline of our proposed LLGS. 1) preprocess images and then initial 3DGS from SfM. 2) the similar rendering pipeline with the 3DGS but with our enhanced color expression M-Color. 3) The M-Color and rendering inputs datails are shown at the right part.}
    \vspace{-14pt}
\end{figure*}
\input{secs/3_method}

\input{secs/4_experiments}
\input{secs/5_conclusion}

\bibliographystyle{IEEEtran}
\bibliography{main}


\end{document}

%% file: secs/0_abstract.tex
\begin{abstract}
3D Gaussian Splatting has shown remarkable capabilities in novel view rendering tasks and exhibits significant potential for multi-view optimization.
However, the original 3D Gaussian Splatting lacks color representation for inputs in low-light environments. Simply using enhanced images as inputs would lead to issues with multi-view consistency, and current single-view enhancement systems rely on pre-trained data, lacking scene generalization. These problems limit the application of 3D Gaussian Splatting in low-light conditions in the field of robotics, including high-fidelity modeling and feature matching. To address these challenges, we propose an unsupervised multi-view stereoscopic system based on Gaussian Splatting, called Low-Light Gaussian Splatting (LLGS). This system aims to enhance images in low-light environments while reconstructing the scene. Our method introduces a decomposable Gaussian representation called M-Color, which separately characterizes color information for targeted enhancement. Furthermore, we propose an unsupervised optimization method with zero-knowledge priors, using direction-based enhancement to ensure multi-view consistency. Experiments conducted on real-world datasets demonstrate that our system outperforms state-of-the-art methods in both low-light enhancement and 3D Gaussian Splatting.


\end{abstract}

%% file: secs/1_introduction.tex
\section{INTRODUCTION}

3D Gaussian Splatting (3DGS)~\cite{kerbl20233d} exhibits impressive performance across robotics, augmented reality (AR), and virtual reality (VR) by representing scene points as Gaussian distributions. These distributions are modeled using specific parameters, including position, color, and volume, and are optimized to minimize errors in the input images.

In comparison to traditional neural radiance fields (NeRF)~\cite{mildenhall2021nerf}, 3DGS effectively manages the complexity of multi-view scenes and achieves real-time rendering with greater efficiency. Consequently, this technology holds significant promise for various robotic applications, such as exploration, search and rescue, and transportation scenarios where 3DGS-based SLAM and planning techniques ~\cite{keetha2024splatam,li2024ngm,matsuki2024gaussian,peng2024rtg,lei2024gaussnav,li2024sgs} can be employed. In the realm of embodied intelligence, the integration of 3DGS for three-dimensional reconstruction enables robots to comprehensively understand scene details ~\cite{bortolon20246dgs}.

\begin{figure}[t]  
  \centering  
  \includegraphics[width=3.4in]{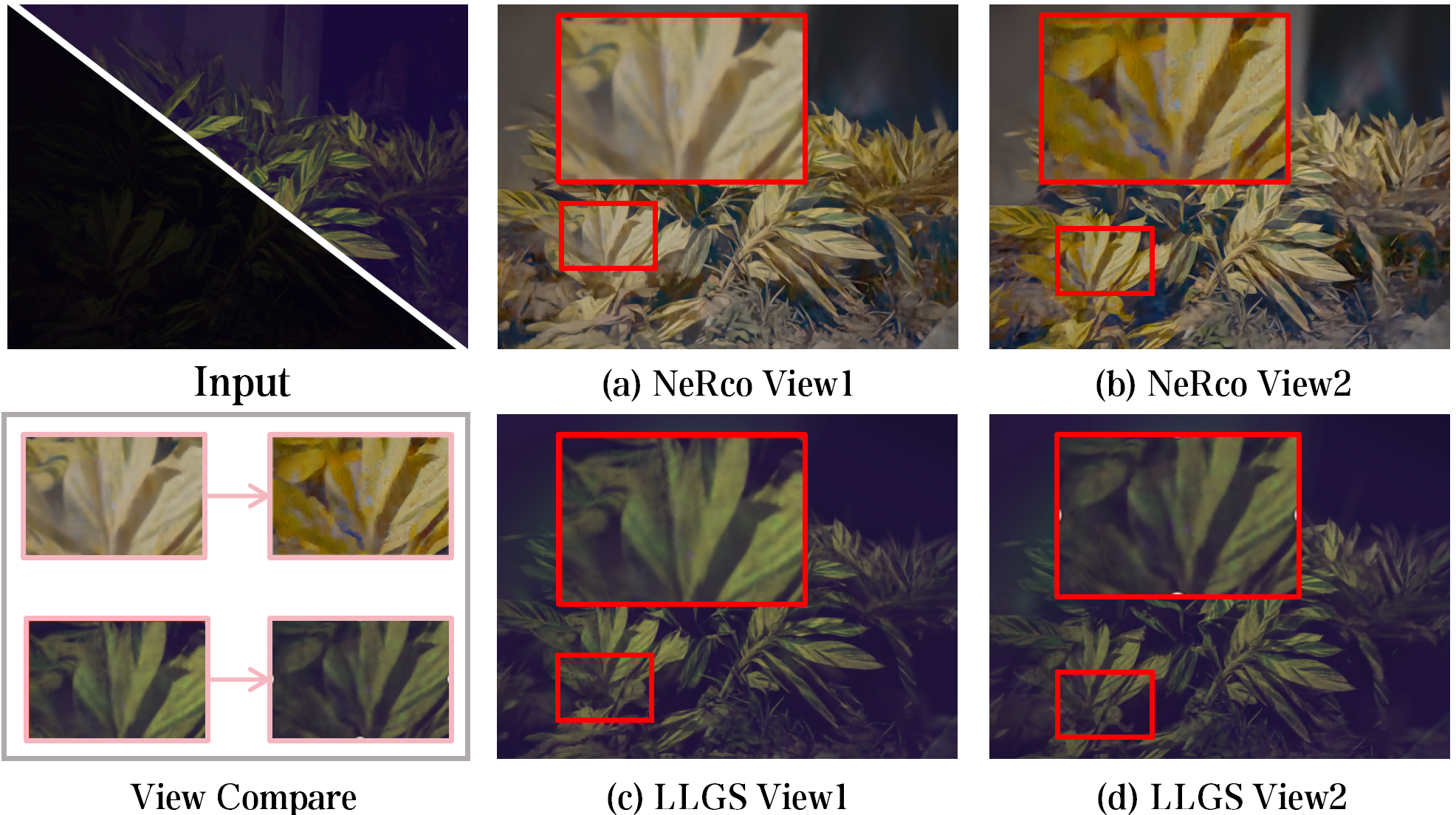}  
  \caption{A comparative between LLGS and the SOTA low-light enhancement model NeRco, using identical multi-view inputs, shows that images (a) and (b) exhibit a error in view consistency post-NeRco processing, while images (c) and (d) underscore the robust view consistency maintained by LLGS, as emphasized in the \textbf{View Compare} image through detailed magnification.}  
  \label{mvs}  
  \vspace{-12pt}
\end{figure}

However, current 3DGS systems are heavily reliant on the visibility of input images, which limits their effectiveness in low-light conditions. In practical robotic applications, challenging lighting environments—such as mines and underwater settings—often result in images with low visibility and significant noise interference. These issues obstruct the reconstruction of geometric details, colors, and textures, complicating the synthesis of high-quality models from new viewpoints.

One straightforward approach is to directly utilize enhanced images ~\cite{jin2023dnf,ma2022toward,wang2022low,liu2023low,yang2023implicit,cai2023retinexformer} for 3DGS reconstruction. However, this method requires extensive data and computational resources for supervised training to achieve accurate enhancement effects, which limits its generalization capabilities in other scenarios. Moreover, existing low-light enhancement models for 2D tasks typically focus on independent single views, failing to consider the need for consistency among multi-view images. This inconsistency can lead to generalization issues in tasks involving real 3D scenes, including 3D reconstruction and feature matching.
As illustrated in Figure \ref{mvs}, inconsistent multi-view synthesis results can lead to blurriness and data interference. Therefore, developing an optimization method that emphasizes multi-view consistency, along with unsupervised approaches to constrain features and relight images, is of significant practical importance.

A noteworthy approach, LLNeRF~\cite{wang2023lighting}, offers an insightful unsupervised method for deconstructing radiance field learning, achieving impressive generalization for low-light enhancement of original input data. It also addresses the consistency issue between multi-view images by jointly optimizing the radiance field using loss signals from all views during the training process. However, methods based on radiance fields encounter challenges, such as the loss of local high-frequency details and extended training times, which hinder their further development.
Moreover, strategies that utilize neural network output position encoding cannot be directly applied to discrete Gaussian points. Consequently, a unique color representation based on 3DGS has become essential to distinguish it from NeRF when addressing low-light scenes.

To address these issues, we propose Low Light Gaussian Splatting (LLGS), the first 3DGS-based unsupervised system for image enhancement and 3D reconstruction of sRGB images. The 3DGS framework exhibits exceptional performance for scenes with multi-view images as inputs, producing processed images with higher quality than the original inputs by effectively reducing noise.
Inspired by Retinex theory~\cite{land1977retinex}, we replace the Spherical Harmonic~\cite{muller2022instant} function-based color representation in the vanilla 3DGS with an decomposable model, M-Color. This model decomposes the color representation of Gaussians into two components: a view-independent component that captures material properties and a view-dependent component that represents illumination. The final color is determined by the interplay of these two components. This novel approach to Gaussian color representation enables better control over illumination effects, enhancing visual features.
We further propose a brand-new unsupervised low-light enhancement method based on Gaussians, utilizing a loss function grounded in the gray world prior and sharpness metrics for Gaussians. Our method optimizes color to better constrain scene representation. By integrating the loss before and after enhancement across all views through an improved joint optimization strategy, we ensure consistency among multi-view images.

Compared to LLNeRF and methods combining State-of-the-Art (SOTA) 2D image enhancement with the vanilla 3DGS approach, our solution not only resolves the former's inability to render in real-time under low-light conditions but also addresses the latter's issues with image consistency.

Our main contributions are as follows:
\begin{itemize}
\item We propose Low Light Gaussian Splatting (LLGS), a 3DGS reconstruction system based on zero-knowledge priors, capable of unsupervised enhancement and reconstruction of normal-light 3D scenes from low-light images while performing real-time rendering. Experimental results on multiple real-world datasets demonstrate that our method outperforms baseline approaches, effectively reconstructing color details and lighting conditions.
\item We introduce an decomposable 3DGS representation, M-Color. This model separates an object's intrinsic color from ambient lighting using the constraints inherent in 3DGS, eliminating the need for ground truth supervision.
\item We develop a novel unsupervised low-light enhancement method based on enhanced Gaussians. This method not only improves lighting and corrects color but also ensures the consistency of multi-view images through an advanced 3DGS joint optimization strategy.
\end{itemize}

%% file: secs/2_relatedwork.tex
\section{RELATED WORK}

\textbf{Neural Radiance Field (NeRF)}:
NeRF uses multi-layer perceptrons (MLPs) to represent scenes and achieve realistic novel view synthesis. The quality of input images is crucial for this process because NeRF relies on these images for color optimization. Several methods have been proposed to train NeRF models from low-quality inputs. For instance, RawNeRF~\cite{mildenhall2022nerf} trains NeRF directly on camera raw images to address issues of low visibility and noise in low-light scenes. HDR-NeRF~\cite{huang2022hdr} synthesizes novel view HDR images from LDR images with varying exposure levels. Similarly, LLNeRF enhances NeRF training using a set of low-light sRGB images by decomposing learned radiance fields. However, the lengthy training times associated with NeRF limit its practicality for real-time rendering applications, making it unsuitable for robotic tasks. Therefore, there is an urgent need for a faster method of 3D reconstruction.

\textbf{3D Gaussian Splatting (3DGS)}:
3DGS fundamentally transforms the backbone structure of NeRF rendering by introducing a Gaussian distribution, enabling rapid real-time rendering while maintaining the accuracy and quality of 3D reconstruction. Recently, several variant models based on 3DGS have been proposed for 3D reconstruction in low-light scenarios. DarkGS~\cite{zhang2024darkgs} creates a 3DGS variant capable of re-illuminating dark environments by modeling the camera-light source system. However, this method relies on accurate camera-light source models in specific environments and does not generalize well to completely dark scenes. Relightable3DGS and GaussianShader~\cite{gao2023relightable, jiang2024gaussianshader} illuminate dark scenes through variant 3DGS reconstruction forms based on normal light source modeling, but their dependence on these models results in lower versatility. HDRGS~\cite{cai2024hdr} leverages HDR information and metadata (e.g., exposure levels) recorded in RAW images for noise reduction. Nevertheless, sRGB images captured in low-light scenarios differ from RAW images; their low dynamic range and low signal-to-noise ratio make them less suitable for 3DGS reconstruction in such environments. Unlike the aforementioned methods, this paper aims to address the challenge of real-time 3D reconstruction using a set of dark sRGB images, converting them into representations under normal lighting conditions. Given the characteristics of low-light images—such as low visibility, low dynamic range, high noise, and color distortion—our method achieves superior enhancement effects.

%% file: secs/3_method.tex
\section{METHOD}
In this section, we introduce our novel pipeline for LLGS.

\subsection{3DGS Preliminaries}
3D Gaussians Splatting compute the color of pixels by representing each feature point in the scene as an explicit 3D Gaussian, which serves as the fundamental rendering component. Each 3D Gaussian is defined by a full 3D covariance matrix $\Sigma_i$ defined in world space centered at point (mean) $\mu_i$ with a scaling factor $ s_i\in R^3$, a rotation quaternion $ q_i \in R^4 $ and additional attributes such as opacity $o_i$ and color feature $f_i$. Mathematically, a 3D Gaussian is represented as:

\begin{equation}
    G(x)=\mathrm{exp}(-\frac{1}{2}(x-\mu)^T{\Sigma}^{-1}(x-\mu))
\end{equation}

Where $x$ is a arbitrary location in the 3D scene, $\Sigma$ represents the covariance matrix, and $\mu$ represents the mean in 3D space. The Gaussian maintains positive semi-definiteness by using a scaling matrix $S$ and a rotation matrix $R$:
\begin{equation}
\Sigma=RSS^TR^T
\end{equation}
In the vanilla 3DGS, the color of the Gaussians is expressed through Spherical Harmonics (SH), controlled by the viewing direction $v_i$ and the learnable spherical harmonics coefficients $h_{i,lm}$, the formula is as follows:
$$
c_i=\sum_{l=0}^{l_{\mathrm{max}}}\sum_{m=-l}^{l}h_{i,lm}Y_{l}^{m}(v_i), \thickspace{ }{\mathrm{Where}}\thickspace{ }\:l_{\mathrm{max}}=3
$$
$Y_{l}^{m}$ represents the spherical harmonic function, $l$ denotes the order of $Y_{l}^{m}$, which determines the basic shape and the complexity of variation of the spherical harmonic function; $m$ denotes the degree of $Y_{l}^{m}$ , which determines the specific shape and directionality of the SH function for a given order. During the rendering process, 3DGS derives the color of a pixel by blending multiple overlapping ordered Gaussians at that pixel's location:
\begin{equation}
C=\sum_{i\in N}c_i\cdot\alpha_i\prod_{j=1}^{i-1}(1-\alpha_j), \thickspace{ }{\mathrm{Where}}\thickspace{ }\alpha_i=G_i^{2D}o_i
\end{equation}

$\alpha_i$ is obtained by multiplying the opacity $o_i$ with the 2D covariance matrix projected from the 3D covariance matrix $\Sigma$ onto the plane. The color $c_i$ is decoded from color features using SH. During implementation, to maintain a meaningful interpretation of the covariance matrix $\Sigma$ throughout the optimization process, it is parameterized as a unit quaternion $q$ and a 3D scaling vector $s$. Overall, a 3D scene is composed of a series of 3D Gaussian points, each of which possesses a specific set of parameters: \{$o_i$, $r_i$, $s_i$, $p_i$, $c_i$\}. These parameters are optimized by Stochastic Gradient Descent (SGD)~\cite{kingma2014adam} and initialized from the Structure from Motion(SfM)~\cite{schonberger2016structure}. \\

\subsection{Decomposable 3DGS Representation M-Color}

Vanilla 3DGS uses a set of anisotropic Spherical Harmonics (SH) to represent the color of Gaussians. 

Retinex theory~\cite{land1977retinex} suggests that our brain interprets the color of objects based on their reflectance properties and the surrounding light conditions. Inspired by this theory, we found it meaningful to decompose Gaussians' color into two-parts.
However, the form of SH is a hybrid model to express the material color with light effects, and it's impossible to decompose it into two part to express the light and material color. 
Therefore, we proposed a decomposable representation: M-Color to represent Gaussians, the core of M-Color is using a neural network $F_{Color}$ to substitute the SH to express the complex color properties. 
The M-Color has the Gaussians' position $p_i$ and camera pose $v_i$ as the input and the $\omega_i$ represent the light and the $m_i$ represent the basic color of the material. The $m_i$ is only affected by $p_i$. Then, by introducing the view direction, the color $c_i$ of a 3D point can be decomposed into a view-independent component $m_i$ and a view-dependent component $\omega_i$ as follows:
\begin{equation}
    c_i = r_i \circ \omega_i
\end{equation}
The M-Color is consist of feature MLP $F_{feature}$, color MLP $F_{color}$, light MLP $F_{light}$ and enhance model $F_{enhance}$. 
The base componet $m$, $\omega_i$ and $m_i$ are described as:
\begin{equation}
    \begin{cases}
    \begin{alignedat}{2}
    f_i &= F_{feature}(p_i;\theta_{F_{feature}}),\\
    \omega_i &= F_{light}(f_i,v_i;\theta_{F_{light}}),\\
    m_i &= F_{color}(f_i;\theta_{F_{color}}),
    \end{alignedat}
    \end{cases}
    \label{eq:dec}
\end{equation}
$\theta_{F_{name}}$ represents the parameters of the corresponding MLP.
Then, we will enhance the $\omega_i$ to get the enhance $\hat{\omega}_i$ as follows: 
\begin{equation}
    \hat{\omega}_i = F_{enhance}(\omega_i)
\end{equation}
and the enhanced color of each Gaussians $\hat{c}_i$ is:
\begin{equation}
    \hat{c}_i = r_i \circ \hat{\omega}_i
\end{equation}
Finally, we rasterize the image as same as the vanilla 3DGS:
\begin{equation}
    I_r=\{C\}
    \thickspace{}\text{where}\thickspace{}\:
    C=\sum_{i\in N}c_i\cdot\alpha_i\prod_{j=1}^{i-1}(1-\alpha_j)
\end{equation}
\begin{equation}
    I_e=\{\hat{C}\}\thickspace{}\text{where}\thickspace{}\:\hat{C}=\sum_{i\in N}\hat{c}_i\cdot\alpha_i\prod_{j=1}^{i-1}(1-\alpha_j)
\end{equation}
Where the $I_r$, $I_e$ represent rendering image without enhancement and rendering image with enhancement, the $C$ and $\hat{C}$ represent the pixels of corresponding image.

\subsection{Gamma Correction for Preprocess}

Vanilla 3DGS encounters significant optimization issues in low-light scenes. The default GS optimization will rapidly prunes 3DGS points due to the greedy approach's tendency to decrease Gaussian opacity, swiftly diminishing their color effects and resulting in nearly black rendered images. With reduced opacities, Gaussian color properties become irrelevant since rendering color is influenced by dot opacity. To mitigate this, we apply the invertible Eq~\ref{equation10} from~\cite{rahman2016adaptive} to enhance the image in high dynamic range, solely to maintain brightness without improving image quality. Gamma correction is employed for this purpose.

\begin{equation}
    I_g(x,y)=AI(x,y)^{\gamma}
    \label{equation10}
\end{equation}
Where $I$ is the input image and the $I_g$ is the image after preprocess. $A$ and $\gamma$ are hyper-parameters.
This way make sure the Gaussians optimization could be successful in such chanllenging cases, and we finally inverse the image to rendering low light image. 

\subsection{Enhance Model and Unsupervised Method}
\subsubsection{Enhance Model} We first use the Enhance MLP $F_{enhance}$ to get the coefficients for enhance model.
\begin{equation}
    \gamma, \mu = F_{enhance}(\omega_i;\theta_{enhance})
    \label{equation11}
\end{equation}

Then we use Eq~\ref{equation12} to enhance the light $\omega$ for each Gaussians. We believe that dynamical gamma correction can be used to enhance $\omega$ under the constraint of rendering RGB values $\hat{C}_r$ of each pixels, 
\begin{equation}
    \widehat{\omega}_{\mathrm{i}}=\phi(\omega_{i})=(\frac{\omega_{i}}{\mu})^{\frac{1}{\gamma_{0}+\gamma}}
    \label{equation12}
\end{equation}
Where $\mu$ and $\gamma$ is a three-dimensional vector, these two coefficients are output of the enhancement network $F_{enhance}$. $\gamma_0$ is a stable value which is used to initialize the non-linear transform. $\mu$ is used to adjust the lighting gain globally, while $\gamma$ is used to correct color distortion by applying small permutations to $\omega$ in the three color channels, all under the constraint of the unsupervised loss functions. 

\begin{figure*}[htbp!]
    \centering
    \includegraphics[width=\linewidth]{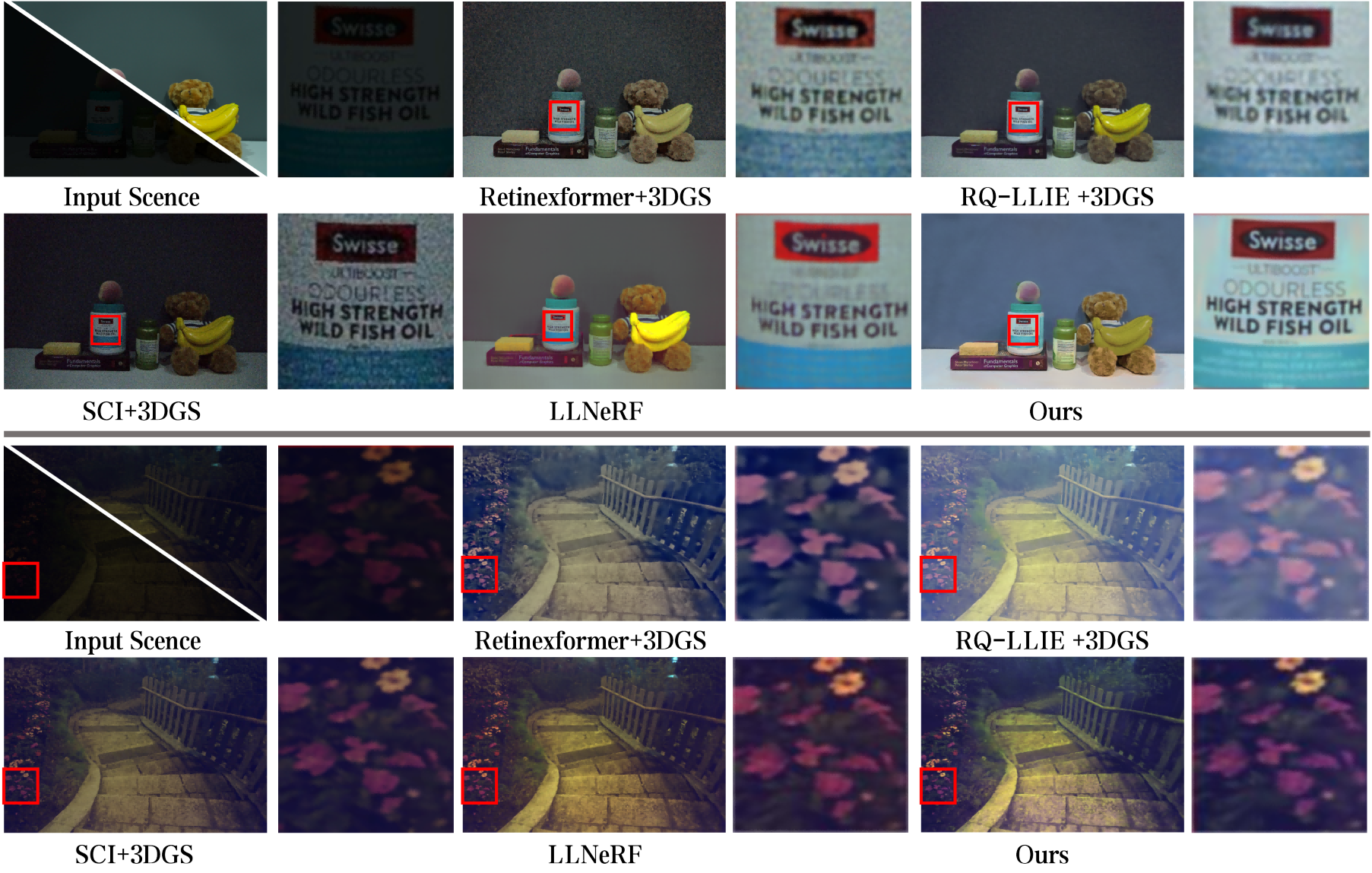}
    \caption{\label{fig:overview}Visual comparison between the SOTA low light 2D image enhancement with vanilla 3DGS and ours. For the \textbf{Input}, the bottom left is the actual input for the two systems, the top right corner is the image in HDR help understand the input information. We compared the results between Retinexformer, RQ-LLIE, SCI, LLNeRF [17], [15], [13], [18]. The results shows that our method demonstrates the best visual effects and the clearest image information.}
    \vspace{-14pt}
\end{figure*}
\subsubsection{Unsupervised Method} We use the following method to optimize our model for unsupervised enhancement.

\textbf{Gray World Assumption Colormetric Supervision}
Gray World Assumption~\cite{lam2005combining,rizzi2002color} suggests that the average gray scale value in an image tends to be consistent. We apply this assumption to control color restoration in a 3D scene by leveraging the process of projecting Gaussians and rendering 2D images using LLGS:
\begin{equation}
    \mathcal{L}_{c}=E[(\hat{C}-e)^{2}]+\lambda_{1}E[\frac{\mathrm{var}(\hat{C})}{\beta_{1}+\mathrm{var}_{c}(r)}]+\lambda_{2}\parallel\gamma\parallel_{2}
\end{equation}
Where $e$, $\lambda_1$, $\beta_1$, and $\beta_2$ are hyperparameters, $\mathrm{var}_c$ represents the channel variance. The first term of $L_c$ is to keep the rendering value at a high level, while the second term corrects colors based on the Gray World Assumption by reducing the variance among the three channels, thus pushing distorted colors towards a natural distribution. Adding a dynamic weight $\gamma$ to the basic color $r$ of the Gaussians can relax the constraints on highly saturated colors.

\textbf{Sharpening Prior Supervision}
For a low light image is easy to loss important geomotry information, we introduce a gradient loss $L_g$ based on the gradient of image, which efficiently keep the edge features. 

\begin{equation}
    \mathcal{L}_g = SSIM(G(I_e), G(I))
\end{equation}

The $I_e$ is the finally enhanced image by 3DGS, $I$ is the input low light image. The structural similarity index measure (SSIM)~\cite{sara2019image, hore2010image} is a function to get the similarity of two images between 0~1. The function $G$ is get the gradient of the image by the Sobel operator~\cite{kanopoulos1988design}, which could be described as follows:

\begin{equation}
    \begin{cases}
    \begin{alignedat}{2}
    g_1(x,y)&=\sum_{k=-1}^1\sum_{l=-1}^1S_1(k,l)f(x+k,y+l),\\
    g_2(x,y)&=\sum_{k=-1}^1\sum_{l=-1}^1S_2(k,l)f(x+k,y+l),\\
    g(x,y)&=\sqrt{g_1^2(x,y)+g_2^2(x,y)},
    \end{alignedat}
    \end{cases}
    \label{eq:dec}
\end{equation}

The $x, y$ is the pixels' index, where $g(x,y)$ on $\hat{I}$ is the new  pixel value indexed by $(x, y)$ on $I$. The $S_1$ and $S_2$ are the Sobel operator denoted as follows: 

\begin{equation}
    S_1 = \begin{bmatrix}+1&0&-1\\+2&0&-2\\+1&0&-1\end{bmatrix}, \space 
    S_2 = \begin{bmatrix}+1&+2&+1\\0&0&0\\-1&-2&-1\end{bmatrix}
\end{equation}

\textbf{Images Supervision}
We also consider the L1 loss $\mathcal{L}_1$ and SSIM loss $\mathcal{L}_\text{D-SSIM}$ between preprocessed image $I_g$ and rendering image $I_r$ as follows: 

\begin{equation}
    \mathcal{L}
    _{image}=(1-\lambda_{SSIM})\mathcal{L}_1+\lambda_{SSIM}\mathcal{L}_\text{D-SSIM}
\end{equation}

%% file: secs/4_experiments.tex
\section{EXPERIMENTS}
In this section, we validate our system on real-world datasets in various environments. We compared our method with SOTA low light image enhancement with vanilla 3DGS and the similar system LLNerf which based on Nerf instead of 3DGS.
   
\subsection{Datasets and comparisons}

\subsubsection{Datasets}

We evaluate the performance of LLGS on the LLNeRF real-world low-light datasets~\cite{wang2023lighting} and the Robotic indoor dataset Replica~\cite{straub2019replica}. The LLNeRF dataset includes 13 scenes for qualitative testing and 3 scenes for quantitative testing, all captured in low-light conditions, both outdoors and indoors, with extremely low average brightness (where the intensity of most pixels is below 50 out of 255).

The Replica dataset comprises 18 highly photo-realistic 3D indoor scene reconstructions at room and building scale, featuring high-resolution high-dynamic-range (HDR) textures. We modify these images to simulate low-light environments, ensuring that the adjusted images maintain an average pixel intensity below 50 out of 255. These extremely low-light datasets present a significant challenge for our task.
\subsubsection{Baseline and comparative methods}
We compared several methods: a baseline method based on 3D Gaussian Splatting (3DGS) with low-light enhancement (LLE) techniques such as Retinexformer~\cite{cai2023retinexformer}, RQ-LLIE~\cite{liu2023low}, SCI~\cite{ma2022toward}, and LLNeRF~\cite{wang2023lighting} based on NeRF. The baseline method, which combines 3DGS with LLE, utilizes the vanilla 3DGS as the backend and is trained on low-light images that have been preprocessed by an LLE frontend. The method based on NeRF is LLNeRF, an enhanced model derived from NeRF that is trained on dark scenes to synthesize new viewpoints.
\subsubsection{Metrics}
To assess the quality of the image enhancement and reconstruction, we adhere to common metrics similar to those used in LLNeRF~\cite{wang2023lighting}, including PSNR and SSIM, by comparing the rendered scenes with the ground truth. Additionally, we evaluate our enhancement results in terms of feature matching to investigate our enhancement's effectiveness in supporting other perception tasks, which is significant in many robotic applications.
\subsubsection{Implementation Details}
We run our system on a PC equipped with an NVIDIA RTX 4090 GPU. We minimize the loss over approximately 30,000 iterations.

\subsection{Experiment Results}

  \begin{figure}[h!]
      \centering
      \includegraphics[width=\linewidth]{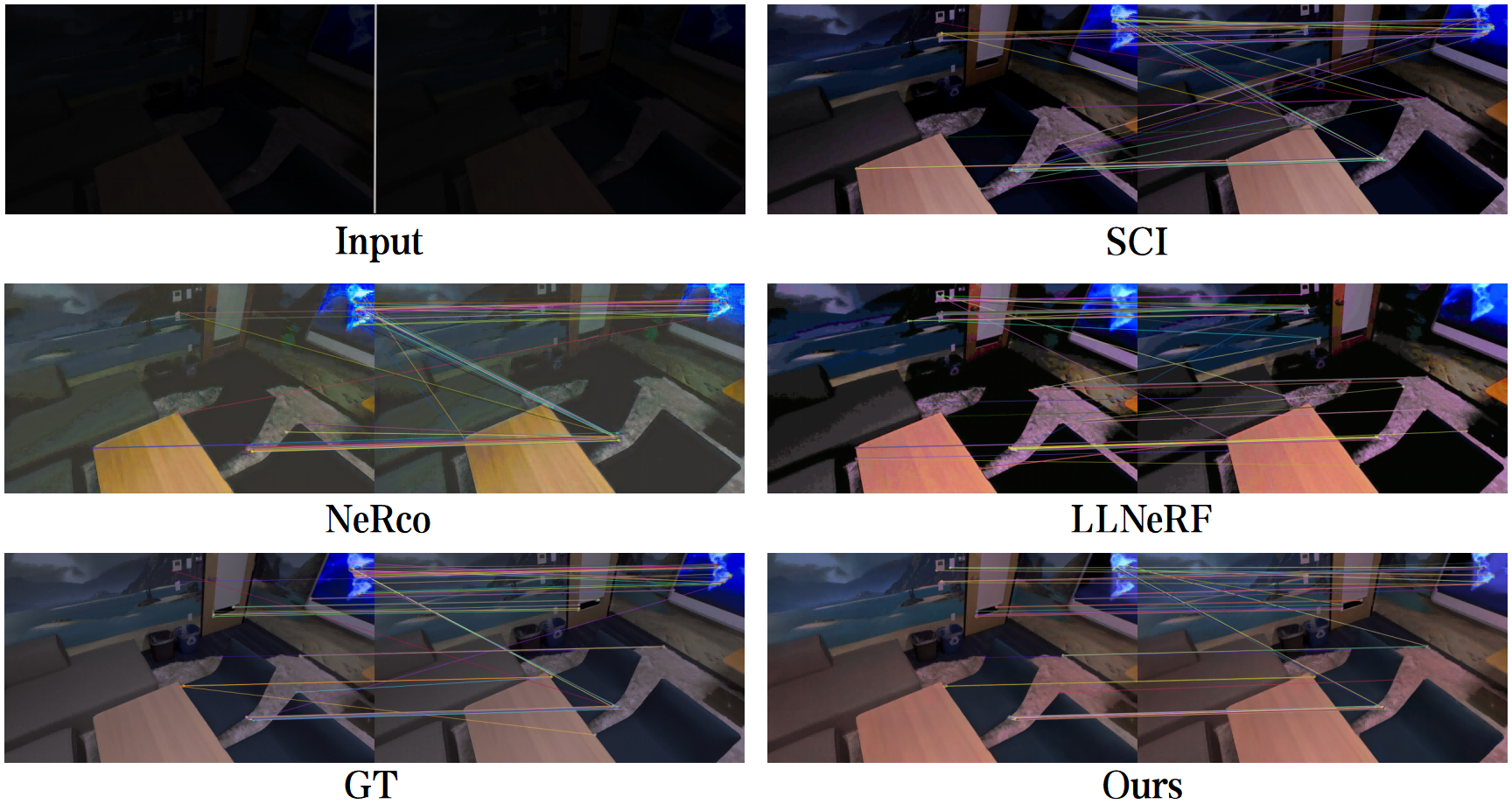}
      \caption{With the input of dark images, our method achieved the highest feature point matching accuracy.}
      \label{fig5}
      \vspace{-15pt}
   \end{figure}

\subsubsection{Quantitative Evaluation}
We compared our method with other SOTA LLE methods using both PSNR and SSIM metrics in LLNeRF datasets. As shown in Table~\ref{tab3}, the results demonstrate that our method outperforms the other SOTA methods. Notably, our method achieves remarkable results in SSIM, indicating that it preserves image information much more effectively.

\subsubsection{Evaluation on Feature Matching}

Our experimental results are presented in Figure~\ref{fig5} and Table~\ref{tab:accuracy}. These results demonstrate that our method can enhance performance in various robotic perception tasks and have the highest Average matching accuracy obtained after enhancing images using three SOTA LLE methods and our method, under the condition of extracting 100 ORB feature points from each image in the Replica dataset. With further advancements, our method has the potential to be deployed in mobile robots to facilitate real-time perception in near-dark environments.

\subsubsection{Runtime}
In Table \ref{tab:time}, we compare the average training time, rendering speed, and GPU memory usage across all datasets. The results indicate that our method offers significant advantages in both speed and GPU requirements.
\vspace{-5pt}

\begin{table}[h!]
\caption{\textbf{Computation \& Runtime \&VRAM.} }
  \centering
  \footnotesize
  \setlength{\tabcolsep}{3pt}
    \begin{tabular}{lccc}
      \toprule
          & Training [min]$\downarrow$ &  Rendering [fps]$\uparrow$ & Memory [GB]$\downarrow$ \\
         \midrule
        {LLNeRF} & 246 & 0.022 & 23.4  \\
        {Ours} & \textbf {47}  & \textbf {136} & \textbf{21.2} \\
       \bottomrule
    \end{tabular}%
    \label{tab:time}
    \vspace{-16pt}
\end{table}

\begin{table}[h!]  
  \caption{\textbf{Feature Matching Accuracy}.}
  \centering  
  \footnotesize  
  \setlength{\tabcolsep}{8pt} 
  \begin{tabular}{lccccc}  
    \toprule  
    Method & SCI & NeRoc & LLNeRF  & GT & Ours \\  
    \midrule  
    Accuracy $\uparrow$ & 74.2\% & 72.1\% & 73.2\% & 76.3\% & \textbf{85.3}\% \\  
    \bottomrule  
  \end{tabular}  
  \label{tab:accuracy}  
  \vspace{-10pt}
\end{table}

    \begin{figure}[h!]
      \centering
      \includegraphics[width=\linewidth]{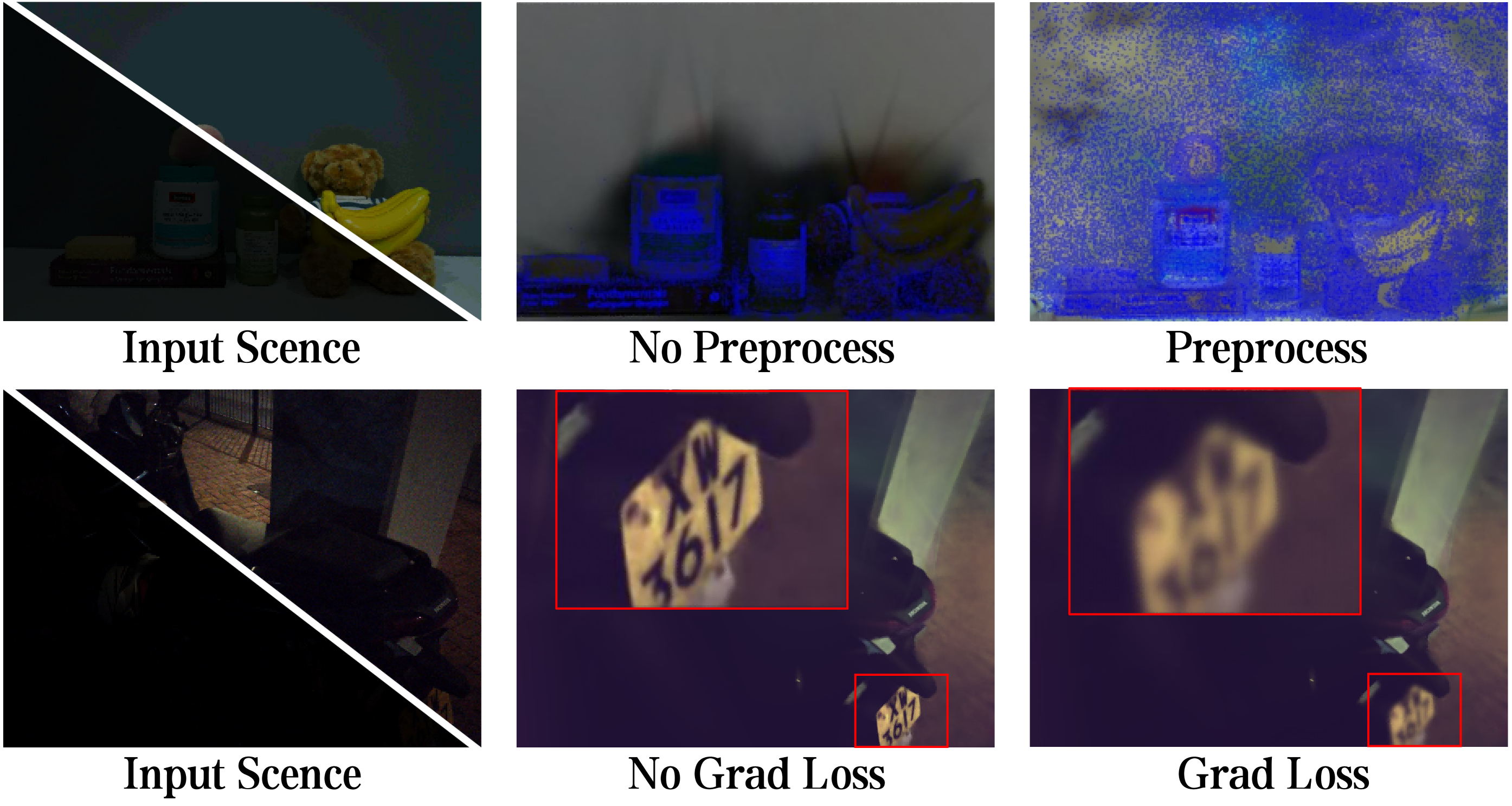}
      \caption{For the \textbf{Input}, the bottom left corner is the actual input for the two systems, the top right corner is the image in HDR. The first line is the ablation study for preprocess. The blue points represent Gaussians. The second line is the ablation study for gradient loss.}
      \label{fig4}
      \vspace{-16pt}
   \end{figure}
   
\begin{table}[tphb]
\caption{\textbf{Quantitative comparison on LLNerf dataset and Replica dataset.}}
\centering
\setlength{\tabcolsep}{9pt}
\begin{tabular}{c|ccc}
\toprule
\textbf{Config} & Scence & PSNR$\uparrow$ & SSIM$\uparrow$\\
\midrule
& Still2 & 18.40 & 0.71 \\
& Still3 & 17.43 & 0.72 \\
Retinex+3DGS & Still4 & 19.12 & 0.82 \\
& office0 & 14.89 & 0.71 \\
& \textbf{Avg} & 17.46 & 0.74 \\
\midrule
& Still2 & 18.99 & 0.74 \\
& Still3 & 17.11 & 0.72 \\
RQ-LLE+3DGS & Still4 & 20.43 & 0.76 \\
& office0 & 16.35 & 0.68 \\
& \textbf{Avg} & 18.22 & 0.73 \\
\midrule
& Still2 & 20.19 & 0.71 \\
& Still3 & 21.29 & 0.73 \\
SCI+3DGS & Still4 & 22.63 & 0.74 \\
& office0 & 19.29 & 0.60 \\
& \textbf{Avg} & 20.85 & 0.72 \\
\midrule
& Still2 & 21.27 & 0.82 \\
& Still3 & 19.85 & 0.71 \\
LLNerf & Still4 & 20.38 & 0.75 \\
& office0 & 18.67 & 0.68 \\
& \textbf{Avg} & 20.04 & 0.74 \\
\midrule
& Still2 & 24.52 & 0.95 \\
& Still3 & 23.76 & 0.94 \\
Ours & Still4 & 24.19 & 0.95 \\
& office0 & 23.22 & 0.92 \\
& \textbf{Avg} & \textbf{23.92} & \textbf{0.94}\\
\bottomrule
\end{tabular}
\label{tab3}
\vspace{-19pt}
\end{table}

\subsection{Ablation Study}
In this section, we highlight the importance of our preprocessing strategy and gradient losses in ensuring the effectiveness of our method.

\subsubsection{Preprocessing}
As illustrated in the Figure~\ref{fig4}, directly inputting low-light images results in significant Gaussian loss, as seen in the bear's head and the wall. This issue leads to more severe problems, including substantial information loss in the enhanced images, which is highly undesirable.

\subsubsection{Gradient Losses}
Figure~\ref{fig4} also shows that omitting gradient loss results in more vague enhanced images. This is exemplified by the difficulty in reading the text in the image.

%% file: secs/5_conclusion.tex
\section{CONCLUSIONS}
We propose a novel unsupervised low-light image enhancement and rendering system, LLGS. This system generates high-quality images with normal lighting to extract clearer information from low-light images in purely dark environments. First, our enhanced Gaussian representation method decomposes Gaussians into a view-dependent component to represent light and a view-independent component to represent the object's color. Seconed, the light and color will be evaluated through 3DGS-based multi-view optimization using our unsupervised method. Finally, we rasterize the enhanced light and color to produce a normally lit image. We conduct experiments to analyze the properties of our method and demonstrate its effectiveness compared to existing methods. Our approach shows potential for future applications in various robotic tasks.